\pdfoutput=1

\documentclass[11pt]{article}

\usepackage[]{naacl2021}

\usepackage{times}
\usepackage{latexsym}

\usepackage[T1]{fontenc}

\usepackage[utf8]{inputenc}

\usepackage{microtype}
\usepackage{booktabs}
\usepackage{CJK}
\usepackage{multirow}
\usepackage{graphicx}
\usepackage{amsmath}
\usepackage{amsfonts}
\usepackage{amssymb}
\usepackage{url}
\usepackage{bm}
\usepackage{tabularx}
\usepackage{color}
\usepackage{pifont}
\usepackage{tikz}

\definecolor{zptu}{RGB}{18, 141, 21}

\title{Robust Dialogue Utterance Rewriting as Sequence Tagging}

\author{Jie Hao$^1$\thanks{~~Work done while J. Hao was interning and L. Wang was working at Tencent AI Lab.}, ~Linfeng Song$^2$, Liwei Wang$^{3*}$, Kun Xu$^2$, Zhaopeng Tu$^2$, \and Dong Yu$^2$ \\
$^1$Florida State University, FL, USA \\
\texttt{haoj8711@gmail.com} \\
$^2$Tencent AI Lab, Bellevue, WA, USA \\
\texttt{\{lfsong,kxkunxu,zptu,dyu\}@tencent.com} \\
$^3$The Chinese University of Hong Kong \\
\texttt{lwwang@cse.cuhk.edu.hk} \\}

\begin{document}
\maketitle
\begin{CJK*}{UTF8}{gbsn}

\begin{abstract}
The task of dialogue rewriting aims to reconstruct the latest dialogue utterance by copying the missing content from the dialogue context. Until now, the existing models for this task suffer from the robustness issue, i.e., performances drop dramatically when testing on a different domain. We address this robustness issue by proposing a novel sequence-tagging-based model so that the search space is significantly reduced, yet the core of this task is still well covered. As a common issue of most tagging models for text generation, the model's outputs may lack fluency. To alleviate this issue, we inject the loss signal from BLEU or GPT-2 under a REINFORCE framework. Experiments show huge improvements of our model over the current state-of-the-art systems on domain transfer.
\end{abstract}

\section{Introduction}

Recent years have witnessed increasing attention in conversation-based tasks, such as conversational question answering \cite{choi2018quac,reddy2019coqa,sun-etal-2019-dream} and dialogue response generation \cite{li2017daily,zhang2018personalizing,wu2019proactive,zhou-etal-2020-kdconv}, mainly due to increasing commercial demands.
However, current models still face tremendous challenges in representing multi-turn conversations.
One main reason is that people tend to use incomplete utterances for brevity, which usually omit (a.k.a. ellipsis) or refer back (a.k.a. coreference) to the concepts that appeared in dialogue contexts.
Specifically, as shown in previous works \cite{su-etal-2019-improving,pan-etal-2019-improving}, ellipsis and coreference exist in more than 70\% of the utterances, which can introduce an extra burden to dialogue models, as they have to figure out those occurrences to understand a conversation correctly.

To tackle the problem mentioned above, the task of dialogue utterance rewriting \cite{su-etal-2019-improving,pan-etal-2019-improving,elgohary2019can} was recently proposed.
The task aims to reconstruct the latest dialogue utterance into a new utterance that is semantically equivalent to the original one and can be understood without referring to the context.
As shown in Table \ref{tab:example}, the incomplete utterance $u_3$ omit ``{\small上海} (Shanghai)'' and refer ``{\small 经常阴天下雨} (always raining)'' with pronoun ``{\small 这样} (this)''.
By explicitly rewriting the dropped information into the latest utterance, the downstream dialogue model only needs to take the last utterance. Thus the burden on long-range reasoning can be largely relieved.

\begin{table} \small
    \centering
    \begin{tabular}{cc}
    \toprule
    Turn & Utterance with Translation \\
    \midrule
    \multirow{2}{*}{$u_1$} & 上海最近天气怎么样？\\
         & (How is the recent weather in Shanghai?) \\
    \midrule
    \multirow{2}{*}{$u_2$} & 最近经常阴天下雨。\\
         & (It is always raining recently.) \\
    \midrule
    \multirow{2}{*}{$u_3$} & 冬天就是\textcolor{blue}{这样}。\\
         & (Winter is like \textcolor{blue}{this}.) \\
    \midrule
    \multirow{2}{*}{$u_3^\prime$} & \textcolor{zptu}{上海}冬天就是\textcolor{blue}{经常阴天下雨}。\\
         & (It is \textcolor{blue}{always raining} in winter \textcolor{zptu}{Shanghai}.) \\
    \bottomrule
    \end{tabular}
    \caption{An example dialogue including the context utterances ($u_1$ and $u_2$), the latest utterance ($u_3$) and the rewritten utterance ($u_3^\prime$).}
    \label{tab:example}
    \vspace{-1.2em}
\end{table}

Most previous efforts \cite{su-etal-2019-improving,pan-etal-2019-improving,elgohary2019can,xu2020semantic} consider this task as a standard text-generation problem, adopting a sequence-to-sequence model with a copy mechanism \cite{gulcehre-etal-2016-pointing,gu-etal-2016-incorporating,see2017get}.
These efforts have demonstrated decent performances on their in-domain test set from the same data source as the training set.
However, they are not robust, as our experiments show that their performances can drop dramatically (by roughly 33 BLEU4 \cite{papineni2002bleu} points and 44 percent of exact match) on another test set created from a different data source (not necessarily from a totally different domain).
We argue that it may not be the best practice to model utterance rewriting as standard text generation.
One main reason is that text generation introduces an over-large search space, while a rewriting output (e.g., $u_3^\prime$ in Table \ref{tab:example}) always keeps the core semantic meaning of its input (e.g., $u_3$).
Besides, exposure bias \cite{wiseman2016sequence} can further exacerbate the problem for test cases that are not similar to the training set, resulting in outputs that convey different semantic meanings from the inputs.

In this paper, we propose a novel solution that treats utterance rewriting as multi-task sequence tagging.
In particular, for each input word, we decide whether to \emph{delete} it or not, and at the same time, we choose what span from the dialogue context need to be inserted to the front of the current word.
In this way, our solution enjoys a far smaller search space than the generation based approaches.

Since our model does not directly take features from the word-to-word interactions of its output utterances, this may cause the lack of fluency.
To encourage more fluent outputs, we propose to inject additional supervisions from two popular metrics, i.e., sentence-level BLEU \cite{chen2014systematic} and the perplexity of a pretrained GPT-2 \cite{radford2019language} model, 
using the framework of ``REINFORCE with a baseline'' \cite{williams1992simple}.
Sentence-level BLEU is computationally efficient, but it requires references and thus may only provide domain-specific knowledge.
Conversely, the perplexity by GPT-2 is reference-free, giving more guidance on open-domain scenarios benefiting from the large-scale pretraining.

Experiments on two dialogue rewriting benchmarks show that our model can give huge improvements (14.6 in BLEU4 score and 18.9 percent of exact match) over the current state-of-the-art model for cross-domain evaluation.
More analysis shows that the outputs of our model keep more semantic information from the inputs.

\section{Related Work}

Initial effort \cite{su-etal-2019-improving,elgohary2019can} treats dialogue utterance rewriting as a standard text generation problem, adopting sequence-to-sequence models with copy mechanism to tackle this problem.
Later work explores task-specific features for additional gains in performance.
For instance, \citet{pan-etal-2019-improving} adopts a pipeline-based method, where all context words that need to be inserted during rewriting are identified in the first step. The second step adopts a pointer generator that takes the outputs of the first step as additional features to produce the output.

\citet{xu2020semantic} train a model of semantic role labeling (SRL) to highlight the core meaning (e.g., who did what to whom) of each input dialogue to prevent their rewriter from violating this information.
To obtain an accurate SRL model on dialogues, they manually annotate SRL information for more than 27,000 dialogue turns, which is time-consuming and costly.
\citet{qian2020incomplete} casts this task into a semantic segmentation problem, a major task in computer vision.
In particular, their model generates a word-level matrix, which contains the operations of substitution and insertion, for each original utterance.
They adopt a heavy model that takes 10 convolution layers in addition to the BERT encoder.
All existing methods only compare in-domain performances under automatic metrics (e.g., BLEU).
We take the first step to study cross-domain robustness, a critical aspect for the usability of this task, and we adopt multiple measures for comprehensive evaluation.
Besides, we propose a novel model based on sequence tagging for solving this task, and our model takes a much smaller search space than the previous models.

\subparagraph{Sequence tagging for text generation}
Given the intrinsic nature of typical text-generation problems (e.g., machine translation), i.e. (1) the number of predictions cannot be determined by inputs, and (2) the candidate space for each prediction is usually very large, sequence tagging is not commonly adopted on text-generation tasks.
Recently, \citet{malmi2019encode} proposed a model based on sequence tagging for sentence fusion and sentence splitting, and they show that their model outperforms a vanilla sequence-to-sequence baseline.
In particular, their model can decide whether to keep or delete each input word and what phrase needs to be inserted in front of it.
As a result, they have to extract a large phrase table from the training data, causing inevitable computation for choosing phrases from the table.
Their approach also faces the domain adaptation issue when their phrase table has limited coverage on a new domain.
Though we also convert our original problem into a multi-task tagging problem, we predict what span to be inserted, avoiding the issues caused by using a phrase table. Besides, we study injecting richer supervision signals to improve the fluency of outputs, which is a common issue for tagging based approaches on text generation, as they do not directly model word-to-word dependencies.
Finally, we are the first to apply sequence tagging on dialogue rewriting, showing much better performances than those BERT-based strong baselines.

\section{Baseline: \textsc{Trans-PG+BERT}}

Our baseline consists of a BERT \cite{devlin2018bert} encoder and a Transformer \cite{vaswani2017attention} decoder with a copy mechanism.
Given input tokens $X=(x_1,\dots,x_N)$ that is the concatenation of the current dialogue context $c=(u_1,\dots,u_{i-1})$ and the latest utterance $u_i$,
the BERT encoder is firstly adopted to represent the input with contextualized embeddings:
\begin{equation} \label{eq:bert_encode}
    \boldsymbol{E} = \boldsymbol{e}_1, \dots, \boldsymbol{e}_N = \text{BERT}(x_1,\dots,x_N) \text{.}
\end{equation}
Next, the Transformer decoder with copy mechanism is adopted to generate a rewriting output $u^\prime=(y_1,\dots,y_M)$ one token at a time:
\begin{align}
    p(y_{t}|y_{<t},X) &= \theta_t p_t^{vocab} + (1-\theta_t) p_t^{attn} \\
    p_t^{attn}, \boldsymbol{s}_t &= \text{TransDecoder}(y_{<t}, \boldsymbol{E}) \\
    p_t^{vocab} &= \text{Softmax}(\text{Linear}(\boldsymbol{s}_t))
\end{align}
where \emph{TransDecoder} is the Transformer decoder that returns the attention probability distribution $p_t^{attn}$ over the encoder states $\boldsymbol{E}$ and the latest decoder state $\boldsymbol{s}_t$ for each step $t$. Following~\newcite{see2017get}, the generation probability $\theta_{t}$ for timestep $t$ is calculated from the weighted sum for the encoder-decoder cross attention distribution and the encoder hidden states.
\begin{equation}
    \theta_t = \sigma(\boldsymbol{w}^{\intercal}\sum_{n\in[1..N]}(p_t^{attn}[n] \cdot \boldsymbol{e}_n))
\end{equation}
where $\boldsymbol{w}$ represents the model parameter.
In this way, 
the 
copy 
mechanism encourages copying words from the input tokens.
The \textsc{Trans-PG} baseline is trained with standard cross-entropy loss:
\begin{equation}
    \mathcal{L}_{gen} = -\sum_{t\in[1..M]}\log p(y_t|y_{<t},X;\boldsymbol{\theta})
\end{equation}
where $\boldsymbol{\theta}$ represents all model parameters.

\section{\textsc{RaST}: Rewriting as Sequence Tagging}

In this section, we describe how to convert the dialogue rewriting task into a multi-task sequence tagging problem.

\begin{figure}
    \centering
    \includegraphics[width=\linewidth]{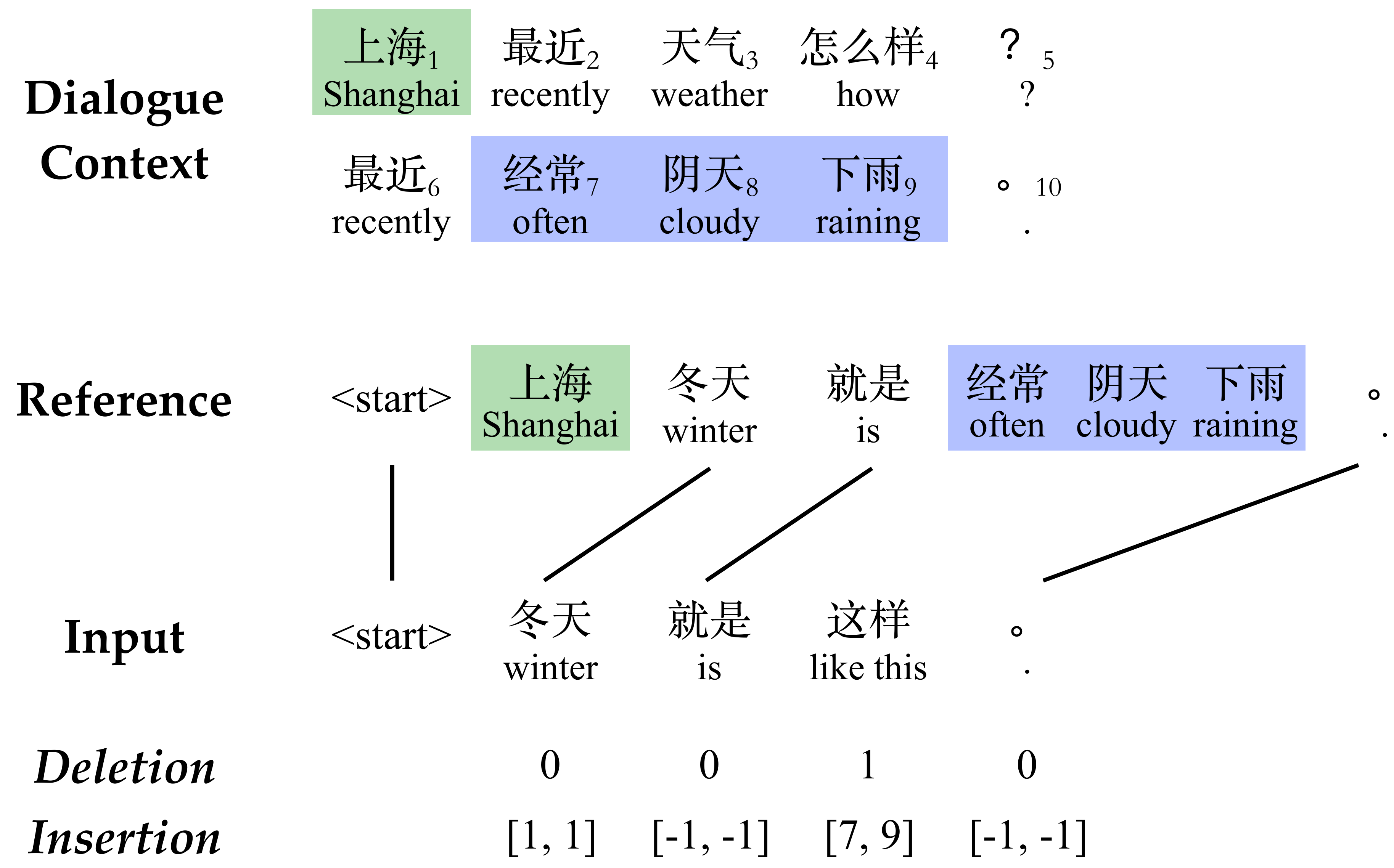}
    \caption{An example of annotating the {\em deletion} and {\em insertion} tags based on the word alignment between the input and reference utterances.}
    \label{fig:annotation_example}
    \vspace{-1.2em}
\end{figure}

\paragraph{Task description}
Our analysis shows that dialogue rewriting mainly handles two linguistic phenomena: coreference and omission.
To recover a coreference, it has to replace a pronoun in the current utterance with the phrase it refers to in the dialogue context. 
To recall an omission, it needs to insert the corresponding phrase into the omission position.
Accordingly, we cast the dialogue rewriting as a sequence tagging task by introducing two types of tags for each word $x_n$:
\begin{itemize}
    \item {\em Deletion} $\in \{0, 1\}$: the word $x_n$ is deleted (i.e. 1) or not (i.e. 0);
    \item {\em Insertion}:[{\em start}, {\em end}]: a phrase ranging the span [{\em start}, {\em end}] in the dialogue context is inserted in front of the word $x_n$. If no phrase is inserted, the span is [-1, -1].
\end{itemize}
Recovering a conference corresponds to the operation \{Deletion:1, Insertion:[{\em start}, {\em end}]\}, and recalling an omission corresponds to the operation \{Deletion:0, Insertion:[{\em start}, {\em end}]\}, where [{\em start}, {\em end}] denotes the corresponding phrase in the dialogue context. For the other words without any change, the operation is \{Deletion:0, Insertion:[-1, -1]\}. Figure~\ref{fig:annotation_example} shows an example, where the word ``{\small 这样} (like this)'' corresponds to a conference, and the word ``{\small 冬天} (winter)''  corresponds to an omission in front of it.\footnote{We use word-based annotations for easier visualization, while character-based annotations are adopted in reality.}

\paragraph{Constructing annotated data}
The gold tags for dialogue utterance rewriting are not naturally available. In response to this problem, we construct the annotated data based on the alignment between the input and reference utterances. Specifically, we employ the \emph{longest common sub-sequence} (LCS)\footnote{\url{https://en.wikipedia.org/wiki/Longest\_common\_subsequence\_problem}} algorithm to generate the word alignments between the input utterance $u_i$ and the reference utterance $u^\prime_i$ for each instance (the black lines in Figure~\ref{fig:annotation_example}).
The LCS algorithm is based on dynamic programming, which takes a time complexity of $\mathcal{O}(|u_i|\times|u^\prime_i|)$. For the words in the reference utterance that are not aligned, we search them from the dialogue context and obtain their span (e.g. the words in color highlighting).

Given the aligned instance, we construct the annotation tags by traversing the alignments in a left-to-right manner and comparing each alignment with its preceding one\footnote{We insert a special flag ``<start>'' at the beginning of both utterances for processing the first alignment.} under the following rules:
\begin{enumerate}
    \item[R1.] If the two alignments are adjacent in both utterances (e.g. ``{\small 就是}~(is)''), there is no change for the current word, which is assigned the tags \{Deletion:0, Insertion:[-1, -1]\}.
    \item[R2.]  If the two alignments are only adjacent in the input utterance (e.g. ``{\small 冬天}~(winter)''), this generally corresponds to an omission. We insert the reference words between the two alignments (i.e. ``{\small 上海}~(Shanghai)'') in front of the current input word. Accordingly, we assign the current word ``{\small 冬天}~(winter)'' the tags \{Deletion:0, Insertion:[1, 1]\}.
    \item[R3.]  If the two alignments are only adjacent in the reference utterance, we simply delete the input words between the two alignments, and assign them the tags \{Deletion:1, Insertion:[-1, -1]\}.
    \item[R4.]  If the two alignments are not adjacent in either utterance (e.g. ``{\small 。}~(.)''), this generally corresponds to a coreference that requires a replacement. We first delete the input words between the two alignments (i.e. ``{\small 这样}~(like this)''), then insert the corresponding target phrase  (i.e. ``{\small 经常~阴天~下雨}~(always raining)'') in front of the left- most deleted input word (i.e. ``{\small 这样}~(like this)''). Accordingly, we assign the left-most deleted word ``{\small 这样}~(like this)'' the tags \{Deletion:1, Insertion:[7, 9]\} (i.e. \emph{replacement}), and assign the current word ``{\small 。}~(.)'' the tags \{Deletion:0, Insertion:[-1, -1]\} (i.e. \emph{no change}). If there exist more than one deleted words, we assign the other words beyond the left-most deleted word the tags \{Deletion:1, Insertion:[-1, -1]\} (i.e. \emph{deletion}).
\end{enumerate}

Both rules \#2 and \#4 require finding phrases from the dialogue context, as highlighted in color in Figure~\ref{fig:annotation_example}. If there are multiple candidates in the dialogue context, we choose the one that is closest to the input utterance to avoid long-range dependency. If no candidate can be found, we consider such instances can not be covered by our approach. In the \textsc{Rewrite} and \textsc{Restoration} datasets used in this work, we respectively found that 6.0\% and 6.5\% instances are not covered, and deleted them from the datasets.

\begin{figure}[t]
    \centering
    \includegraphics[width=0.48\textwidth]{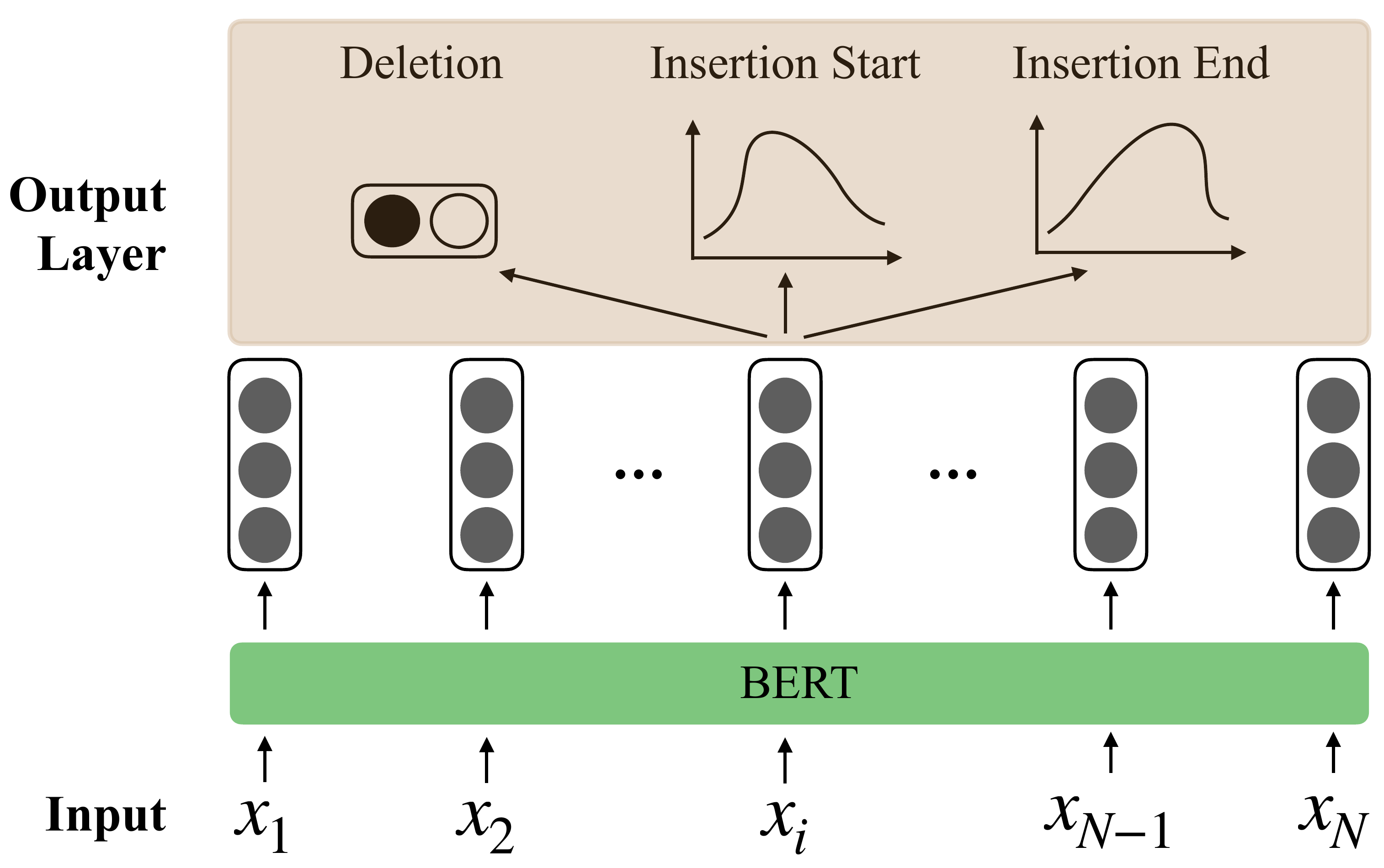}
    \caption{Our model architecture. For fair comparison, it takes the same encoder architecture as the baseline.}
    \label{figmodel_arch}
    \vspace{-1.2em}
\end{figure}

\section{Approach}

\subsection{Model Architecture}

Figure \ref{figmodel_arch} shows the architecture of our model.
For a fair comparison, it takes the same BERT-based encoder (Equation \ref{eq:bert_encode}) as the baseline to represent each input.
For simplicity, we directly apply classifiers to predict the corresponding tags for each input word.
In particular, to determine whether each word $x_n$ in the current utterance $u_i$ should be kept or deleted, we use a binary classifier:
\begin{eqnarray} \label{eq:loss1}
   p(d_{n}|X, n) = \text{Softmax}(\boldsymbol{W}_{d}\boldsymbol{e}_{n}+\boldsymbol{b}_{d})
\end{eqnarray}
where $\boldsymbol{W}_{d}$ and $\boldsymbol{b}_{d}$ are learnable parameters, $d_{n}$ is the binary classification result, and $\boldsymbol{e}_{n}$ is the BERT embedding for $x_n$.


Moreover, we cast span prediction as machine reading comprehension (MRC)~\cite{rajpurkar2016squad}, where a predicted span corresponds to an MRC target answer.
For each input token $x_n\in u_i$, we follow the previous work on MRC to predict the start position  $s^{st}_{n}$ and end position $s^{ed}_{n}$ for the target span $s_n$, performing separate self-attention mechanisms for them:
\begin{align}  \label{eq:loss2}
   p(s^{st}_{n}|X, n) &= \text{Attn}_{start} (\boldsymbol{E}, \boldsymbol{e}_{n}) \\  \label{eq:loss3}
   p(s^{ed}_{n}|X, n) &= \text{Attn}_{end} (\boldsymbol{E}, \boldsymbol{e}_{n})
\end{align}
where $\text{Attn}_{start}$ and $\text{Attn}_{end}$ are the self-attention layers for predicting the start and end positions of a span.
We use the standard additive attention mechanism~\cite{bahdanau2014neural} to perform the attention function.
The probability for the whole span $s_n$ is:
\begin{equation} \label{eq:loss4}
    p(s_{n}|X, n) = p(s^{st}_{n}|X, n) p(s^{ed}_{n}|X, n)
\end{equation}
where $s^{st}_{n}$ is no greater than $s^{ed}_{n}$.

Given an example of ($c$, $u_i$) pair and $X=[c;u_i]$, the overall loss function for the multi-task sequence labeling is defined as the standard cross-entropy loss over gold tags:
\begin{equation}
\begin{split}
   \mathcal{L}_{tagging} = - \sum_{x_n\in u_i} \Big(\log p(d_{n}|X, n) + \\ \log p(s_{n}|X, n) \Big)
\end{split}
\end{equation}
where the terms are defined in Equation \ref{eq:loss1} and \ref{eq:loss4}.


\subsection{Enhancing Fluency with Additional Supervision}

By converting dialogue utterance rewriting into a sequence tagging task, our model enjoys better efficiency and lower search space.
However, a potential side effect is that our outputs may lack fluency, because our approach does not directly model word-to-word dependencies.

We explore sentence-level BLEU \cite{chen2014systematic} and GPT-2 \cite{radford2019language} as additional training signal to improve the fluency of our generated outputs,
adopting the framework of ``REINFORCE with a baseline'' to inject these supervision signals.
For more detail, we first generate two candidate sentences: one is by \emph{sampling} the tags at each position of the input utterance according to the distributions in Equation \ref{eq:loss1} and \ref{eq:loss4}, the other is by greedily choosing the \emph{model-considered best} tags.
Next, the RL objective for sample ($c$, $u_i$) is calculated by: 
\begin{equation}
   \mathcal{L}_{rl} = (r(\hat{u}_i^{g}, u_i)-r(\hat{u}_i^{s}, u_i))\log p(\hat{u}_i^{s}|X)
\end{equation}
where $\hat{u}_i^{s}$ and $\hat{u}_i^{g}$ represents the two candidate sentences by sampling and greedy ``argmax'', respectively.
$r(\cdot,\cdot)$ is the reward function, which can correspond to either sentence-level BLEU or the perplexity by the GPT-2 model.
Finally, we follow previous work by combining this additional loss with the tagging loss:
\begin{eqnarray}
   \mathcal{L} = (1-\lambda) \mathcal{L}_{tagging} + \lambda \mathcal{L}_{rl},
\end{eqnarray}
where the $\lambda$ is a constant weighting factor that is empirically set to $0.5$.



\section{Experiments}

We study the robustness of our tagging-based model on two benchmarks for dialogue rewriting.

\subsection{Setup}

\begin{table} \small
    \centering
    \begin{tabular}{c|c|c}
        Dataset & \textsc{Rewrite} & \textsc{Restoration} \\
    \hline
    Train/dev/test size & 18k/1k/1k & 194k/5k/5k\\
    No. turns & 3 & 6 \\
    No. con. tokens ($\mu$, $\sigma$) & (18.63, 8.50) & (38.36, 11.71) \\ 
    \end{tabular}
    \caption{Statistics of two datasets, where ``No. turns'' denotes number of turns for each example, and ``No. con. tokens ($\mu$, $\sigma$)'' denotes the mean and standard deviation for number of tokens in the context.}
    \label{tab:data}
    \vspace{-1.2em}
\end{table}

\begin{table*}[t]
\begin{center}
\small
\renewcommand\arraystretch{1.1}
\begin{tabular}{c|l||c|c||c|c|c|c|c|c}
      \# & {\bf Model}   & {\bf BLEU1}   & {\bf  BLEU2}  &  {\bf BLEU3 } & {\bf BLEU4} & {\bf R1} & {\bf R2} & {\bf R-L}  &{\bf EM}\\
      \hline
      \multicolumn{10}{c}{{\em Evaluation on the \textsc{Rewrite} Test Set}} \\
      \hline
      1& \textsc{Trans-PG}+BERT & 88.1 & 86.5 & 84.9 & 82.3 & 90.2 & 84.1 & 91.0 & 51.3 \\ 
      2& \textsc{Csrl}~\cite{xu2020semantic} & 89.0 & 87.5 & 85.6 & 83.5 & 89.9 & 81.5 & 87.5 & 47.4  \\
      3& \textsc{Run}~\cite{qian2020incomplete} & \textbf{93.5} & \textbf{90.9} & 88.2 & 85.5 & \textbf{95.8} & \textbf{90.3} & 91.3 & \textbf{65.1} \\
      \hline 
      4& \textsc{RaST} & 90.6 & 90.2 & \textbf{89.3}  &  \textbf{88.2} &  94.0 &  88.9  & \textbf{91.5} & 64.3 \\
      5& \textsc{RaST}+RL-BLEU & 89.9 & 89.6 & 88.7 & 87.7 & 93.7 & 88.7 & 91.2 & 64.4 \\
      6& \textsc{RaST}+RL-GPT2 & 89.2 & 88.8 & 87.9 & 86.9 & 93.5 & 88.2 & 90.7 & 63.0  \\
      \hline 
      \multicolumn{10}{c}{{\em Evaluation on the \textsc{Restoration} Test Set}} \\
      \hline 
      7 &  \textsc{Trans-PG}+BERT & 65.8 & 61.4 & 58.5 & 55.3 & 66.9 & 55.1 & 69.8 & 7.4 \\ 
      8 & \textsc{Csrl}~\cite{xu2020semantic} &  70.0 & 66.1 & 63.2 & 60.1 & 67.4 & 55.9 & 67.6 & 8.6 \\
      9 &\textsc{Run}~\cite{qian2020incomplete} & 79.3 & 74.6 & 70.2 & 65.7 & 81.2 & 69.6 & 76.5 & 12.9  \\
      \hline 
      10 & \textsc{RaST} & 84.8 & 83.2 & 81.4 & 79.3 & 82.9 & 74.1 & 78.8 & 24.5  \\
      11 & \textsc{RaST}+RL-BLEU & 84.4 & 82.9 & 81.2 & 79.1 & 83.3 & 74.7 & 79.3 & 26.8  \\
      12 & \textsc{RaST}+RL-GPT2 & \textbf{85.2} & \textbf{83.8} & \textbf{82.2} & \textbf{80.3} & \textbf{84.3} & \textbf{76.0} & \textbf{80.5} & \textbf{31.8}  \\ 
\end{tabular}
\caption{Test results of all comparing models trained on the \textsc{Rewrite} dataset.}
\label{tab:acl_data_train}
\vspace{-1.5em}
\end{center}
\end{table*}

\subparagraph{Datasets}

We conduct experiments on two popular dialogue rewriting datasets: \textsc{Rewrite} \cite{su-etal-2019-improving} and \textsc{Restoration} \cite{pan-etal-2019-improving}.
Both datasets are created by first crawling multi-turn dialogues from popular Chinese social media platforms, before asking human annotators to generate the rewriting result for the last turn of each dialogue.
Table \ref{tab:data} lists some statistics of both datasets.
In addition to the difference on data scale (20K vs 204K), it shows additional disparities on other aspects.
For instance, the number of turns for \textsc{Restoration} is twice as much as \textsc{Rewrite}, and the dialogue context size for \textsc{Restoration} is larger and more variant than that of \textsc{Rewrite}.
Besides, around 40\% instances in \textsc{Restoration} do not require any changes for rewriting, while all instances require rewriting for \textsc{Rewrite}.
These differences make transferring between the two datasets a good test bed for evaluating model robustness.

\subparagraph{Model settings}
We implement the baseline and our model on top of a BERT-base model~\cite{devlin2018bert}, and
we use Adam~\cite{kingma:adam} as the optimizer, setting the learning rate to $3^{-5}$ as determined by a development experiment.
For the reinforcement learning stage, we respectively use the sentence-level BLEU score with ``Smoothing 3''~\cite{chen2014systematic} or the perplexity score based on a Chinese GPT-2 model trained on massive dialogues \cite{zhang-etal-2020-dialogpt}\footnote{https://github.com/yangjianxin1/GPT2-chitchat} as the reward function.
It is worth noting that the GPT-2 model is not fine-tuned during the reinforcement learning stage.

\subparagraph{Comparing models}
In addition to the \textsc{Trans-PG+BERT} baseline, we compare our approach with several state-of-the-art dialogue rewriting models that are also based on BERT.
\textsc{Csrl}~\cite{xu2020semantic} leverages additional information on conversational semantic role labeling (CSRL) to enhance BERT representation, extra human efforts are required on CSRL annotation.
\textsc{Run}~\cite{qian2020incomplete} treats this problem as semantic segmentation by predicting a word-level edit matrix for the input utterance. 
For fair comparison, we either run their released model or ask the authors to generate their outputs on our data.

\vspace{0.2em}
\textbf{Evaluation}~~
We use both automatic metrics and human evaluations to compare our proposed model with other approaches.
For the automatic metrics, we follow previous work to use BLEU \cite{papineni2002bleu}, ROUGE \cite{lin2004rouge} and the percent of sentence-level exact match (EM score).

\subsection{Main Results}
\label{sec:main}

\subparagraph{Training on \textsc{Rewrite}}
Table~\ref{tab:acl_data_train} shows the results when all comparing models are trained on the \textsc{Rewrite} dataset, before evaluating on the in-domain \textsc{Rewrite} and the out-of-domain \textsc{Restoration} test data.
On the \textsc{Rewrite} test set, our tagging-based models (Rows 4-6) are much better than the \textsc{Trans-PG+BERT} baseline, and they can get comparable performances with the previous state-of-the-art model \textsc{Run}.
\textsc{Run} usually get high numbers on BLEU1 without consistent improvements on higher-order BLEU scores.
Our observation shows that it tends to insert context words into wrong places, hurting the number of matches regarding higher $n$-gram.

Among our tagging-based models, we find that injecting additional training signal (Row 5-6) slightly hurts the in-domain performance.
For \textsc{RaST+RL-GPT2}, the reason can be that optimizing with the perplexity of GPT-2 will dilute the main signal on the in-domain task.
But, it can enjoy better generality on other domains.
Comparatively, \textsc{RaST+RL-BLEU} reports closer numbers to the baseline than \textsc{RaST+RL-GPT2}, 
because the reward by sentence-level BLEU optimizes towards a similar direction with the main signal.
In general, both types of rewards can improve the fluency of model outputs, especially on other domains.


\begin{table*}[t]
\begin{center}
\small
\renewcommand\arraystretch{1.1}
\begin{tabular}{c|l||c|c||c|c|c|c|c|c}
      \# & {\bf Model}   & {\bf BLEU1}   & {\bf  BLEU2}  &  {\bf BLEU3 } & {\bf BLEU4} & {\bf R1} & {\bf R2} & {\bf R-L}  &{\bf EM}\\
      \hline
      \multicolumn{10}{c}{{\em Evaluation on the \textsc{Restoration} Test Set}} \\
      \hline
      1 & \textsc{Trans-PG}+BERT & 88.0 &  87.0 & 85.9 & 84.4 & 90.1 & 84.5 & 89.8 & 49.0  \\ 
      2 & \textsc{Csrl}~\cite{xu2020semantic} & 90.6 & \textbf{89.7} & \textbf{88.6} & \textbf{87.2} & 91.1 & 85.0 & \textbf{90.0} & 49.1 \\
      3 & \textsc{Run}~\cite{qian2020incomplete} & \textbf{92.0} & 89.1 & 86.4 & 83.6 & \textbf{92.1} & \textbf{85.4} & 89.5 & \textbf{49.3}  \\
      \hline 
      4 & \textsc{RaST} & 89.7 & 88.8 & 87.6 & 86.1 & 91.1 & 84.2 & 87.8 & 48.7 \\
      5 & \textsc{RaST}+RL-BLEU & 90.4 & \textbf{89.6} & \textbf{88.5} & \textbf{87.0} & 91.2 & 84.3 & 87.9 & 48.8 \\
      6 & \textsc{RaST}+RL-GPT2 & 89.7 & 88.9 & 87.7 & 86.2 & 90.9 & 84.0 & 87.6 & 47.8  \\
      \hline 
      \multicolumn{10}{c}{{\em Evaluation on the \textsc{Rewrite} Test Set}} \\
      \hline 
      7 &  \textsc{Trans-PG}+BERT & 76.6 & 74.9 & 73.0 & 70.8 & 79.8 & 70.7 & 79.7 & 25.7 \\
      8 & \textsc{Csrl}~\cite{xu2020semantic} &  75.5 & 73.8 & 71.8 & 69.6 & 80.0 & 69.9 & 79.0 & 23.2 \\
      9 &\textsc{Run}~\cite{qian2020incomplete} &  80.6 & 76.0 & 71.1 & 65.9 & 84.5 & 73.6 & 80.6 & 27.2 \\
      \hline 
      10 & \textsc{RaST} & 81.2 & 80.0 & 78.2 & 75.9 & \textbf{84.9} & \textbf{75.2} & \textbf{81.1} & 28.5 \\
      11 & \textsc{RaST}+RL-BLEU & 80.9 & 79.6 & 77.8 & 75.5 & 84.8 & 75.1 & 80.7 & \textbf{29.6}  \\
      12 & \textsc{RaST}+RL-GPT2 & \textbf{82.4} & \textbf{81.0} & \textbf{79.1} & \textbf{76.7} & \textbf{85.0} & 74.8 & 80.8 & \textbf{29.4} \\ 
\end{tabular}
\caption{Test results of all comparing models trained on the \textsc{Restoration} dataset.}
\label{tab:emnlp_data_train}
\vspace{-1.5em}
\end{center}
\end{table*}

When we look at the evaluation results on the out-of-domain \textsc{Restoration} test set, all the compared baseline models (Rows 7-9) get much worse performances than the in-domain situation, where the drops are 27, 23 and 20 BLEU4 points for \textsc{Trans-PG+BERT}, \textsc{Csrl} and \textsc{Run}, respectively.
On the other hand, the performance drops of our tagging models (Rows 10-12) are much smaller, where the gaps are 9, 8.6 and 6.6 BLEU4 points for \textsc{RaST}, \textsc{RaST+RL-BLEU} and \textsc{RaST+RL-GPT2}, respectively.
This comparison demonstrates the superiority of our models on cross-domain robustness.
As the result, \textsc{RaST+RL-GPT2} achieves much higher numbers than previous state-of-the-art models across all metrics.


\vspace{0.2em}
\textbf{Training on \textsc{Restoration}}~~
To further verify the above conclusions, we also conduct experiments in the opposite direction by training all models on the \textsc{Restoration} dataset, and the results are shown in the Table~\ref{tab:emnlp_data_train}.
Similarly, Our models achieve comparable performances with the current state-of-the-art systems and the \textsc{Trans-PG+BERT} baseline on the in-domain test set.
This time, the performance gaps for the three comparison systems (Rows 1-3) are 13.6, 17.6 and 17.7 BLEU4 points, which are lower than the gaps in Table \ref{tab:acl_data_train}.
This is mainly due to the large size of the \textsc{Restoration} training data, which is 10 times larger than \textsc{Rewrite}.
Nevertheless, our models still manage to outperform all comparing systems with a descent margin.

In addition to the conclusions above, we find some other interesting facts.
The advantage of \textsc{Run} on BLEU1 still does not benefit high order situations.
This is consistent with the situation in Table \ref{tab:acl_data_train}, which may infer a systematic defect of \textsc{Run} that inserts words into wrong contexts.
Comparing the two types of supervisions, GPT-2 is better on the out-of-domain test set, while sentence-level BLEU is better on the in-domain test set.
This is intuitive as GPT-2 is reference-free and can provide general guidance, while the reference-based BLEU score is more domain-specific.


\subsection{Human Evaluation}

In addition to these automatic metrics, we also conduct a human evaluation for each system pair to further compare their rewriting quality, and we focus on the \emph{out-of-domain} scenario.
Specifically, we first take the models trained on \textsc{Rewrite} to decode 500 randomly selected test examples from \textsc{Restoration}, before asking 3 graders to choose a winner for each pair of rewriting outputs.
The evaluation criteria is based on fluency and adequacy, where the adequacy mainly considers two aspects: 1) what percent of meaning is retained; 2) how many coreference and omission situations are recovered.
All graders agree on 88.8\% cases.

\begin{table}[t]
\begin{center}
\small
\renewcommand\arraystretch{1.1}
\begin{tabular}{l|c|c|c}
      {\bf Model}   &  {\bf Win} & {\bf Tie} & {\bf Loss} \\
      \hline 
      Ours v.s. Trans-PG+BERT & 23.0\%  & 63.2\% &  13.8\% \\
      Ours v.s. \textsc{Csrl} & 28.4\% & 57.0\%  & 14.6\% \\
      Ours v.s. \textsc{Run} & 23.6\% & 64.0\% & 12.4\%   \\
\end{tabular}
\caption{Human evaluation on 500 randomly selected test samples from \textsc{Restoration}. ``Ours'' denotes the \textsc{RaST}+RL-GPT2 model.}
\label{tab:human}
\vspace{-1.5em}
\end{center}
\end{table}

\begin{table*}[t]
\begin{center}
\small
\renewcommand\arraystretch{1.1}
\begin{tabular}{l|c|c|c|c|c|c}
      \multirow{2}{*}{\bf Model}  & \multicolumn{3}{c|}{\bf Transfer to \textsc{Restoration}} &  \multicolumn{3}{c}{\bf Transfer to \textsc{Rewrite}} \\
       \cline{2-7}
         &  {Precision} & {Recall} & {F-score} & {Precision} & {Recall} & {F-score}  \\
      \hline 
      \textsc{Trans-PG}+BERT & 25.74  &  24.72 &  25.22 &  41.16 & 36.81 & 38.86 \\
      \textsc{Csrl} & 24.67 & 26.13  & 25.38 & 40.36 & 35.97 & 38.04 \\
      \textsc{Run} & 26.52 &  27.62 &  27.06  & 41.09 & 37.78 & 39.36 \\
      \textsc{RaST}+RL-GPT2 & \textbf{37.40} & \textbf{36.90} & \textbf{37.15} & \textbf{41.85} & \textbf{39.41} & \textbf{40.59} 
\end{tabular}
\vspace{-0.5em}
\caption{Evaluation based on semantic role labeling for rewrites. ``Transfer to \textsc{Restoration}'' denotes models trained on the \textsc{Rewrite} training data and tested on \textsc{Restoration} test data, and vice versa.}
\label{tab:srl}
\vspace{-1.0em}
\end{center}
\end{table*}

\begin{table*}[t]
\begin{center}
\footnotesize
\renewcommand\arraystretch{1.0}
\begin{tabular}{l|l|l|l}
    & Case1  & Case2  & Case3\\
    \hline
    Contexts & U1:  {\footnotesize 我意见很大} & U1: {\footnotesize 帮我找一下西安到商洛的顺风车} & U1: {\footnotesize 你帮我考雅思} \\
    ~~(Translation) & {\scriptsize (I have a lot of complaints)} & {\scriptsize (Can you help me find a free ride from Xi'an to Shangluo)} & {\scriptsize (Please help me on IELTS)}  \\
    & U2: {\footnotesize 有意见保留} &  U2: {\footnotesize 哪的} & U2: {\footnotesize 雅思第一项考什么} \\
    & {\scriptsize (keep it yourself if there's any)} & {\scriptsize (where is it)} & {\scriptsize (What is tested first for IELTS)}  \\
    \hline 
    Current utterance & U3: {\footnotesize 不想保留} & U3: {\footnotesize 能不能找到} & U3: {\footnotesize 考口语啊} \\
    ~~(Translation) & {\scriptsize (don't want to keep it myself)} & {\scriptsize (can you find any)} & {\scriptsize (it's oral test)}  \\
    \hline 
    Reference & {\footnotesize 不想保留意见} & {\footnotesize 能不能找到西安到商洛的顺风车} & {\footnotesize 雅思第一项考口语啊} \\
    ~~(Translation) & {\scriptsize (don't want to keep the} & {\scriptsize (can you find any free ride} & {\scriptsize (it's oral test for IELTS)}  \\
    & {\scriptsize complaints myself)} & {\scriptsize from Xi'an to Shangluo)} & {\scriptsize }  \\
    \hline 
    \textsc{Trans-PG}+BERT & {\footnotesize 不想保留} &  {\footnotesize 能不能找到商洛的顺风车} & {\footnotesize \textcolor{red}{考口语考口语}啊} \\
    \textsc{Run} &  {\footnotesize 意见不想\textcolor{red}{意}保留} & {\footnotesize 能不能找到商洛的顺风车} & {\footnotesize 雅思第一项考口语啊} \\
     \textsc{RaST}+RL-GPT2 & {\footnotesize 意见不想保留} & {\footnotesize 能不能找到西安到商洛的顺风车}  & {\footnotesize 雅思第一项考口语啊} \\
\end{tabular}
\vspace{-0.5em}
\caption{Case Study.}
\label{tab:case_study}
\vspace{-1.5em}
\end{center}
\end{table*}

As shown in Table~\ref{tab:human}, the number of winning cases for \textsc{RaST+RL-GPT2} is much more than the number of losing cases when comparing with any other system.
This further confirms the effectiveness of our model.
Many losing cases are due to the lack of fluency caused by object fronting, while most of this type of situations are understandable by human.
Some examples are discussed in Section \ref{sec:case}.
Our model loses the least number of samples against \textsc{Run}, because \textsc{Run} also lack fluency due to improper context word insertion (as mentioned in Section \ref{sec:main}).

\subsection{Evaluating with Semantic Role Labeling}

We additionally compare our model with the baselines by evaluating the semantic role labeling (SRL) results on their outputs.
By doing so, we can approximate the semantic correctness of the rewriting outputs.
Specifically, we choose a state-of-the-art SRL system~\cite{che2020n}
to annotate rewriting outputs and references.
Next, the precision, recall and F1 scores regarding the SRL annotations are calculated for every comparing system.

Table \ref{tab:srl} lists the performances regarding SRL on both scenarios of domain transfer, where our model reports the highest numbers on both directions.
This indicates that our model also makes improvement regarding the core semantic meaning.
Overall, the relative differences on SRL is consistent with the differences under automatic metrics (Table \ref{tab:acl_data_train}, \ref{tab:emnlp_data_train}) and human evaluation (Table \ref{tab:human}).




\subsection{Case Study}
\label{sec:case}

Table \ref{tab:case_study} gives 3 test examples that indicate the representative situations we find.
The first example illustrates the cases when \textsc{Run} inserts context words (e.g. ``{\small \textcolor{red}{意}}'') into wrong places.
This hurts the fluency and high-order BLEU score as mentioned in Section \ref{sec:main}.
The third example shows the situations when the \textsc{Trans-PG+BERT} baseline messes up by word repeating (e.g. ``{\small \textcolor{red}{考口语考口语}}'').
This is a common situation for generation-based models, especially on unseen data samples.
Conversely, this situation rarely happens to our model, as it is based on sequence tagging.
Lastly, the second example corresponds to the situation of referring to a complex concept (e.g. ``{\small 西安到商洛的顺风车} (a free ride from Xi'an to Shangluo)'').
For these cases, it is easier for our model to get the correct span.
This is because our model directly predicts the span boundaries, thus it has a smaller search space than other previous approaches, like generating the concept word by word.


\subsection{Evaluation on Uncovered Examples}
\label{sec:uncover}

As mentioned earlier, a few examples may not be covered by our model, which treats rewriting as sequence tagging.
To get a more comprehensive evaluation, we further compare the baseline and our model on the combination of the \emph{uncovered}  test examples for both \textsc{Rewrite} and \textsc{Restoration} datasets.
Table \ref{tab:uncover} shows the performances in the out-of-domain setting.
Comparing with the baseline, our model is comparable on precision (indicated by BLEU4) and is better at content recall (indicated by Rough-L).
Regarding EM, our model achieves none, because all testing examples are \emph{not covered} by our model.
On the other hand, the EM score for the baseline is also close to 0.0.
Our investigation finds that most of these examples are very challenging, we list representative examples in the Appendix.

\begin{table}[t]
\begin{center}
\small
\begin{tabular}{l|c|c|c}
      {Model} & {BLEU4}  & {R-L} &  {EM} \\
      \hline 
      \textsc{Trans-PG}+BERT & 56.6 & 67.6 & 0.6 \\
      \textsc{RaST}+RL-GPT2 & 56.4 & 70.7 & 0.0
\end{tabular}
\caption{Performances on the combination of the \emph{uncovered} test examples from both datasets.}
\label{tab:uncover}
\vspace{-1.5em}
\end{center}
\end{table}


\section{Conclusion}
In this paper, we addressed the robustness issue of domain transfer for dialogue utterance rewriting, which is crucial for its usability on downstream applications.
We proposed a novel tagging-based approach that takes a much less search space than the existing methods on this task, and we introduced additional supervision (e.g. by GPT-2) to improve the fluency of model outputs.
Experiments show that our model improves the previous state-of-the-art system by a large margin regarding both automatic metrics and human evaluation.

\bibliography{custom}
\bibliographystyle{acl_natbib}

\clearpage
\appendix

\section{Examples for Uncovered Cases}
\label{sec:appendix}

\begin{table}[th!]
\begin{center}
\small
\renewcommand\arraystretch{1.1}
\begin{tabular}{l|l}
    \multicolumn{2}{l}{\textbf{\#1}} \\
    \hline
    Contexts & U1:  流浪地球有看吗  \\
    (Translation) & {\scriptsize (Have you seen ``The Wandering Earth'')} \\
    & U2: 总感觉流浪地球写成短篇浪费了太精彩了 \\
    & {\scriptsize (I fill that it's a waste to make a short film out of ``The Wandering Earth'', the novel is awesome)} \\
    \hline 
    Current utterance & U3: 其实导演剪辑版有两小时半 \\
    (Translation) & {\scriptsize (In fact the director's cut version is 2.5 hour long)} \\
    \hline 
    Reference & 其实导演剪辑版的流浪地球有两小时半  \\
    (Translation) & {\scriptsize (In fact the director's cut version of ``The Wandering Earth'' is 2.5 hour long)} \\
    \hline 
    \textsc{Trans-PG}+BERT & 其实导演剪辑版有两小时半  \\
    \textsc{RaST}+RL-GPT2 & 其实流浪地球导演剪辑版有两小时半 \\
    \hline
    \multicolumn{2}{l}{\textbf{\#2}} \\
    \hline 
    Contexts & U1:  有图有真相征男友 \\
    (Translation) & {\scriptsize (Seeking for boyfriend. I'm with my real personal photographs)} \\
    & U2: 上盘真相先 \\
    (Translation) & {\scriptsize (Upload your photographs first)} \\
    & U3: 头像就是 \\
    (Translation) & {\scriptsize (See my profile photo)} \\
    & U4: 看不清 \\
    (Translation) & {\scriptsize (It's unclear)} \\
    \hline 
    Current Utterance & U5: 那肿么办 \\
    (Translation) & {\scriptsize (What can I do)} \\
    \hline 
    Reference & 头像看不清那肿么办 \\
    (Translation) & {\scriptsize (What can I do if it's unclear)} \\
    \hline 
    \textsc{Trans-PG}+BERT & 像看不清男友肿  \\
    \textsc{RaST}+RL-GPT2 & 肿么办 \\
    \hline
    \multicolumn{2}{l}{\textbf{\#3}} \\
    \hline 
    Contexts & U1:  一个人的孤单 \\
    (Translation) & {\scriptsize (A person's loneliness)} \\
    & U2: 一个人吃饭看书写信到处走走停停 \\
    (Translation) & {\scriptsize (I'm single and everyday I taste cuisine, read books, write letters, and travel around)} \\
    & U3: 这也太悠闲了点我还是有正经事要考虑的 \\
    (Translation) & {\scriptsize (You are so relaxed, I'd do something regular)} \\
    & U4: 我这是写的文艺了点除了上班就这些 \\
    (Translation) & {\scriptsize (I described that in an artistic way, I have a regular job)} \\
    \hline 
    Current Utterance & U5: 那你日子过得也不错哦 \\
    (Translation) & {\scriptsize (You have a very good life)} \\
    \hline 
    Reference & 除了上班就看书写信那你日子过得也不错哦 \\
    (Translation) & {\scriptsize (You have a very good life, if you have time to read and write after work)} \\
    \hline 
    \textsc{Trans-PG}+BERT & 那你的孤孤单单子过得也不错哦  \\
    \textsc{RaST}+RL-GPT2 & 那你日子过得也不错哦 \\
    \hline
\end{tabular}
\caption{Case study for the uncovered examples.}
\label{tab:case_study_uncovered}
\end{center}
\end{table}

\end{CJK*}
\end{document}